\pdfoutput=1

\documentclass[11pt]{article}

\usepackage{ACL2023}

\usepackage{times}
\usepackage{latexsym}

\usepackage[T1]{fontenc}

\usepackage[utf8]{inputenc}

\usepackage{microtype}

\usepackage{inconsolata}
\usepackage{amsfonts}
\usepackage{amsmath}
\usepackage{graphicx}
\usepackage{subcaption}
\usepackage{amsmath}
\usepackage{amsthm}
\usepackage{booktabs}
\usepackage{algorithm}
\usepackage{algorithmic}
\usepackage{multirow}
\usepackage{multicol}
\usepackage{mathptmx}

\renewrobustcmd{\bfseries}{\fontseries{b}\selectfont}
\renewrobustcmd{\boldmath}{}
\newrobustcmd{\B}{\bfseries}

\title{Object-Attribute Binding in Text-to-Image Generation: Evaluation and Control}

\author{Maria Mihaela Trusca\textsuperscript{1,*},  Wolf Nuyts\textsuperscript{1,*}, Jonathan Thomm\textsuperscript{2}, Robert Hönig\textsuperscript{2}, \\ \textbf{Thomas Hofmann\textsuperscript{2}, Tinne Tuytelaars\textsuperscript{1}, Marie-Francine Moens\textsuperscript{1}} \\
\textsuperscript{1} KU Leuven, Department of Computer Science, \textsuperscript{2} ETH Zurich, Department of Computer Science \\
\texttt{\{mariamihaela.trusca, wolf.nuyts, tinne.tuytelaars, sien.moens\}}@kuleuven.be, \\ \texttt{\{jonathan.homm, wolf.nuyts, robert.hoenig, thomas.hofmann\}}@inf.ethz.ch \\
* Equal contribution
}

\begin{document}
\maketitle
\begin{abstract}
Current diffusion models create photorealistic images given a text prompt as input but struggle to correctly bind attributes mentioned in the text to the right objects in the image. This is evidenced by our novel image-graph alignment model called EPViT (Edge Prediction Vision Transformer) for the evaluation of image-text alignment. To alleviate the above problem, we propose focused cross-attention (FCA) that controls the visual attention maps by syntactic constraints found in the input sentence. Additionally, the syntax structure of the prompt helps to disentangle the multimodal CLIP embeddings that are commonly used in T2I generation. The resulting DisCLIP embeddings and FCA are 
easily 
integrated in state-of-the-art diffusion models without additional training of these models. We show substantial improvements in T2I generation and especially its attribute-object binding on several datasets.\footnote{Code and data will be made available upon acceptance.}
   
\end{abstract}

\section{Introduction}
\label{sec:intro}

Text-to-image synthesis (T2I) refers to the process of generating visual content based on textual input. The goal is to create realistic images that correctly align with the provided textual descriptions. Recent advances are mainly 
attributed to the introduction of large-scale diffusion models trained on millions of text-image pairs, such as DALL-E 2 \cite{Ramesh2022_DALLE}, GLIDE \cite{Nichol2021Glide}, Imagen \cite{saharia2022imagen} and open-source models such as Stable Diffusion \cite{Rombach2022_ldm} and DeepFloyd IF \cite{saharia2022imagen}.
While these models generate high-quality photorealistic images, their performance drops when multiple objects are mentioned in the textual prompts, especially because of the incorrect binding of attributes to objects
\cite{Rassin2022_doubles_survey, Ramesh2022_DALLE, saharia2022imagen, Chefer2023AttentAndExcite}. 
They often bind objects with their most common attributes, for instance, when prompted with ``a golden car and a red watch'', the model might generate a golden watch and a red car. Another problem is that they spread an attribute's influence over multiple objects in the image (attribute leakage). For example, ``a golden ingot and fish'' generates a goldfish and golden ingot \cite{Rassin2022_doubles_survey}.

\begin{figure}[t!]
    \centering
   \includegraphics[width=0.48\textwidth]{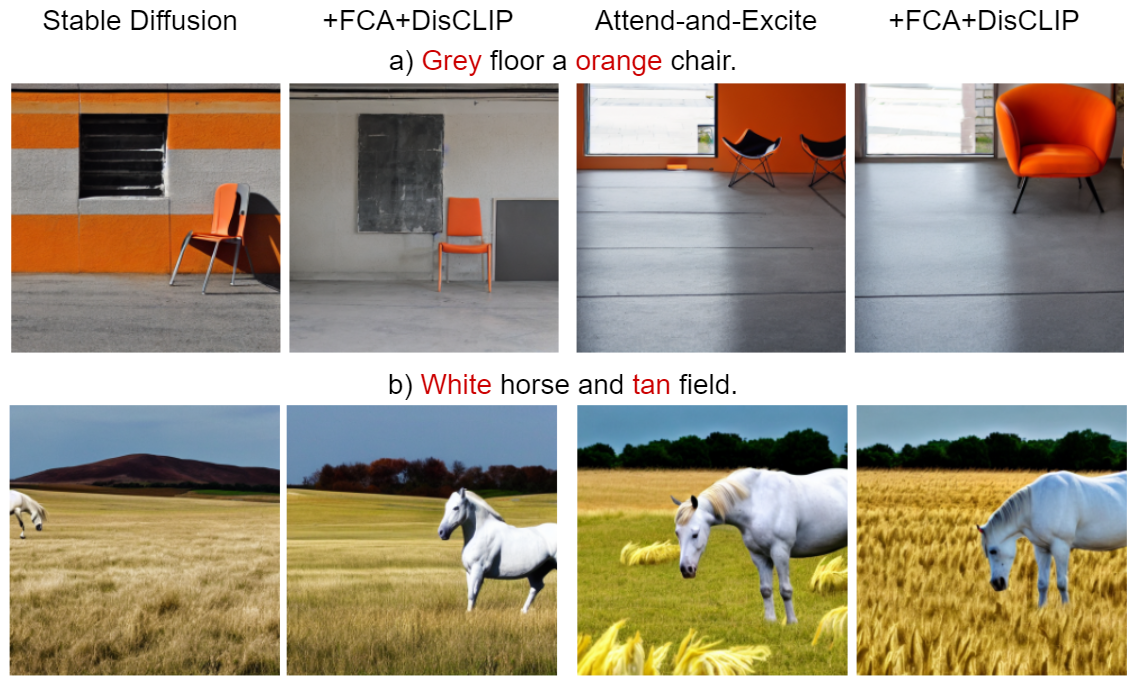}
    \caption{Examples of integrating focused cross-attention (FCA) and disentangled CLIP embeddings (DisCLIP) into
    Stable Diffusion and Attend-and-Excite resulting in (a) a decrease of attribute leakage and b) improved object-attribute binding.}
    \label{fig:teaser}
\end{figure}

As a \textit{first contribution}, we introduce an evaluation set-up to assess a correct image-text alignment. Currently, T2I evaluation is mainly performed based on the CLIP score \cite{Hessel2021CLIPscore}, which assesses the overall semantic match between the textual description and the generated image but fails to check the correct binding of attributes to objects in case of complex multi-object prompts. In contrast, our evaluation set-up, which consists of prompts and confusing adversarial ones, assesses the alignment by reporting how accurate a separately trained evaluation model is able to select from which prompt an image is generated. 
For assessing a fine-grained alignment between the text prompt and the generated image, we propose a novel ViT-based \cite{Dosovitskiy2020ViT} prediction model, called EPViT, that relies on an image graph and that significantly outperforms CLIP in the assessment of image-text alignment. The EPViT predictions form the basis for computing the EPViT accuracy, which is well suited for evaluating object-attribute binding.

\begin{figure}[t!]
    \centering
   \includegraphics[width=0.48\textwidth]{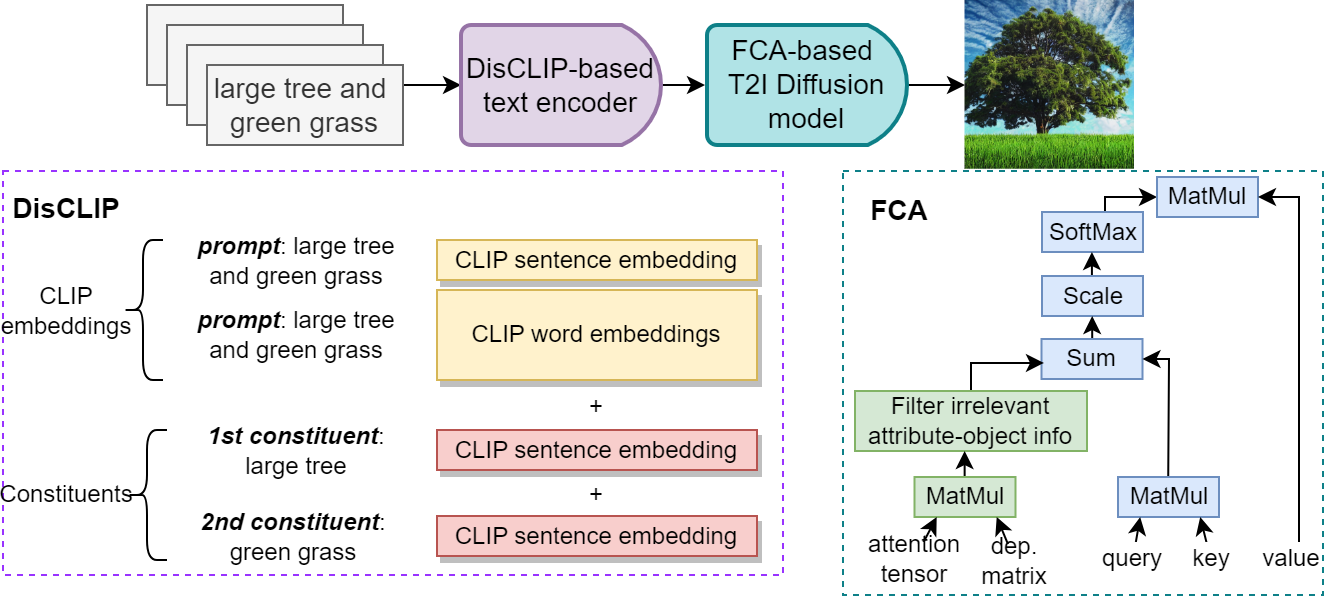}
    \caption{Integration of FCA and DisCLIP in a diffusion-based T2I model is straightforward. While DisCLIP encodes the input prompt, the cross-attention of the diffusion model is easily replaced by its FCA variant.}
    \label{fig:overview}
\end{figure}

As a \textit{second contribution}, we propose two components that leverage the syntactic structure of the text prompt and integrate them into diffusion-based training-free models. Training-free refers to the use of diffusion models trained on large-scale data as commonly done in the literature
instead of retraining them. A first component, focused cross-attention (FCA), constrains the visual attention maps by the syntactic structure found in the input sentence. FCA restricts the attention of attributes to the same spatial location to which their corresponding objects attend. 
Second, the syntactic structure helps to disentangle the multimodal prompt embeddings that are commonly used in T2I generation. We propose a novel \textit{disentangled CLIP} (DisCLIP) encoding that relies on a syntax parser to generate a constituency tree from a sentence.
and 
that mitigates the entanglement issues observed with CLIP encodings. 
Both FCA and DisCLIP result in improved attribute binding and in a decrease in attribute leakage as seen in Figure \ref{fig:teaser}. An extra benefit is their easy integration in any diffusion-based T2I model (Figure \ref{fig:overview}).

As a \textit{third contribution} we conduct an in-depth human and quantitative evaluation of the proposed the EPViT evaluation score and of the T2I generation, the latter showing the superiority of the proposed methods that leverage the syntactic structure of the language prompt compared to current state-of-the-art methods. 

\section{Related Work}

\paragraph{Evaluation of T2I Generation} Most prior works use the Inception Score (IS) \cite{Salimans2016IS} and Frechet Inception Distance (FID) \cite{Heusel2017FID} when evaluating image fidelity (i.e., congruence with the text prompt) of T2I models, while both metrics have proven to imperfectly reproduce human preferences \cite{Heusel2017FID, Parmer2022FIDrobustness}.
The CLIP score \cite{Hessel2021CLIPscore} is also popular, notwithstanding its poor performance at counting and often incorrect binding of attributes \cite{Ramesh2022_DALLE, Radford2021clip}. It uses CLIP embeddings for the computation of the semantic similarity between a text and an image \citep{Park2021R-prec}. 
We propose a new alignment evaluation set-up, and its EPViT accuracy score significantly outperforms the CLIP score for the assessment of a correct alignment between the text prompt and the generated image, when it comes to object-attribute binding. 

\paragraph{T2I Generation Using Diffusion Models}
The introduction of diffusion models \cite{Nichol2021Glide} in combination with classifier-free guidance \cite{ho2022classifierfree} has led to a large leap in image quality. 
\citet{Rassin2022_doubles_survey} demonstrate that T2I models have problems with concept leakage and homonym duplication. \citet{petsiuk2022human_eval} show the low performance of these models on sentences with multiple objects, attributes and relationships. To mitigate the above problems, many models have improved the spatial control of image generation by leveraging spatial constraints in the form of scene layouts to guide the diffusion process. This guidance is commonly realized by relying on additional resources that detect objects and their bounding boxes or segments in the image and by exploiting the object label and region associations of the attention maps, e.g., \cite{liu2022compositional, hertz2022_prompttoprompt,  Avrahami_2023_CVPR, chen_WACV_2024, Li_2023_CVPR, Xie_2023_ICCV, Yang_2023_CVPR, wang2024instancediffusion}, or by the use of sketches as is done by ControlNet \cite{Zhang_2023_ICCV}. In this work we do not use such additional resources but rely on the syntactic structure of the text prompt to realize the guidance.   
\citet{feng2023structguidance} adapts the Stable Diffusion backbone to attend to multiple encodings representing syntactic constituent of the text prompt. Similar to this work, which we use as a baseline, we leverage the syntax of the text prompt but we explicitly exploit syntactic dependencies to bind attributes to objects, leading to better T2I generation. Attend-and-Excite \cite{Chefer2023AttentAndExcite} improve the cross-attention between objects mentioned in the text prompt and the image embeddings showing that their method is especially suited to generate multiple objects. SynGen \cite{rassin2023linguistic} syntactically analyzes the prompt and uses this information in appropriate loss functions that enhance the similarity between the attention maps of objects and their attributes while increasing the distance between these attention maps and those of other words in the prompt.
We show that we can integrate the proposed FCA and DisCLIP encoding in a seamless way in state-of-the-art T2I generation baselines among which are Attend-and Excite and SynGen, and improve their results.

\section{Preliminaries}

\paragraph{Cross-Attention in Diffusion Models} 
The U-Net of diffusion models \cite{saharia2022imagen, Rombach2022_ldm, Nichol2021Glide} use cross-attention layers to condition a denoising network $\epsilon_\theta$ on a text prompt $y$.
A common implementation of this cross-attention uses query (here encoded image), key and value (here encoded text) attention of \citet{Vaswani2017attention} to calculate the cross-attention maps $A^l_t \in \mathbb{R}^{h \times w, n}$ for each layer $l$ and timestep $t$ of the denoising process:
\begin{equation}
    Q^l_t = x_t^l W_Q^l, \; K^l = y W_K^l, \; V^l = y W_V^l
\end{equation}
\begin{equation}
    A^l_t = softmax(\dfrac{Q^l_t(K^l)^{\textnormal{T}}}{\sqrt{d}}), \; f^l_t = A^l_t V^l
\end{equation}
where $W_Q^l$ represents a linear layer transforming $x_t^l$ into the queries $Q^l_t \in \mathbb{R}^{h \times w, d}$, where $d$ denotes the feature dimension. Similarly, $W_K^l$ and $W_V^l$ transform $y$ into keys $K^l \in \mathbb{R}^{n, d}$ and values $V^l \in \mathbb{R}^{n, d}$. $f^l_t\in \mathbb{R}^{h\times w, d}$ represents the output features of the cross-attention layer. Details on 
diffusion models are presented in Appendix \ref{diffusion}.

\section{New Evaluation Framework}
\subsection{Difficult Adversarial Attributes (DAA-200)}
The proposed evaluation set-up assesses the image-text alignment of T2I models. This set-up is based on a new dataset mined from the Visual Genome \cite{Krishna2016VisualGenome} dataset, called Difficult Adversarial Attributes (DAA-200). 
DAA-200 uses the image-graph pairs of Visual Genome to obtain 100 quadruplets of the form \{attribute 1, object 1, attribute 2, object 2\}. 
These quadruplets can be represented in a simple graph with the two objects as nodes and one attribute for each node. From each graph, an adversarial graph is generated by swapping the attributes of both objects. For each of the 200 graphs a sentence of the form $\langle attribute 1 \rangle \langle object 1\rangle and \langle attribute 2 \rangle \langle object 2 \rangle$.'' is generated. To ensure that we have difficult adversarial examples, for DAA-200 we picked examples from Visual Genome where both objects occur multiple times in Visual Genome with each of the attributes. We use this dataset to assess the image-text alignment of a T2I generation model 
and report the accuracy of selecting the text prompt (original or adversarial) on which the generation of the image was conditioned. 



\subsection{Edge Prediction Vision Transformer (EPViT)}
We create an evaluation model whose purpose is to assess whether relationships and attributes are satisfied in an image. This is realized by augmenting and finetuning a Vision Transformer model (ViT) \cite{Dosovitskiy2020ViT}, initialized from CLIP \cite{Radford2021clip}, to take two object nodes of a graph as additional input, classify relationships between the nodes and classify their attributes. Our model essentially predicts the edges of a graph, hence its name Edge Prediction Vision Transformer (EPViT). We hypothesize that by explicitly learning to predict each attribute and relationship, our model becomes better at grounding attribute and relationship dependencies in the image than generic image-text alignment models like CLIP. 

To input the object embeddings to the 
Vision Transformer while preserving the ViT pretrained structure, EPViT uses a method inspired by ControlNet \cite{zhang2023controlNet}. This method makes use of the special input token $ce$ that a Vision Transformer uses in addition to the input image.  $ce$ is a learned embedding, called class embedding. A zero convolution is used to add the object embeddings $obj_1$ and $obj_2$ to $ce$, which results in $ce^*$:
\begin{equation}
    \label{eq:zero_conv}
    ce^* = ce + \alpha * (W*\textsc{concat}(obj_1,obj_2)+b) + \beta
\end{equation}
where $\alpha, \beta \in \mathbb{R}$, are the parameters of a 1$\times$1 convolution, initialized with $0$. $W,b$ are the learnable weights of a linear layer, used to transform the object embeddings into the dimension of $ce$. With $\alpha,\beta=0$ 
this initialization is the original ViT. During finetuning, gradients adapt $\alpha,\beta$ to be nonzero, and the model gradually focuses on the additional input without destroying the original model.

In addition to conditioning the ViT model on two object nodes (encoded with a CLIP text encoder to obtain object embeddings), five linear classification heads are added on top of the ViT model: one relationship classifier head, two attribute classifier heads and two object heads. The two object heads only exist to strengthen the gradients, that is, to ensure that the objects' input information is not ignored. 
The relationship and object classifiers use a single cross-entropy loss, while the attribute classifiers use 
independent binary classifiers using a binary cross-entropy loss (because an object can have multiple attributes). 

\paragraph{Visual Genome Training Details}
Similar to \cite{zhang2017vsgClean, Li2020vsgClean}, we train with the top-most-occurring 100 labels for relationships and attributes, and the top 200 labels for objects after converting them to lowercase and removing trailing spaces. Additionally, we remove all the samples of the DAA-200 dataset from the training set. 
If a sample has no relationship between objects, then the relationship loss of this sample is obscured. Similarly, we obscure the attribute loss for the objects that have no attributes. Note that predicting no relationship/attribute is not a desirable solution because, while there is no relationship/attribute annotation for that object, there is often one present in the image (the dataset is sparse and noisy).
In the experiments below, we focus on datasets that contain 
attribute dependencies, however, the training and design of EPViT allows it to be used on more complex graphs that include relationships between objects.

\paragraph{Using EPViT as a Prediction Model}
For a given graph and image, we average the predicted log-likelihoods, obtained with EPViT, over each relationship and attribute present in the graph. This average is used to assess how well the image represents the graph. 
We calculate the {\bf EPVIT accuracy} as the percentage of generated images for which the graph of the correct sentence obtained a higher average than the graph of the adversarial sentence. 

\section{Methods to Improve Object-Attribute Binding in T2I Generation}
We propose two training-free methods to improve the text conditioning of diffusion models. The first method called focused cross-attention (FCA), leverages a syntactic parse of the text prompt to restrict the attention of an attribute to regions where the corresponding object is active. This method integrates well with diffusion models based on large language encoders trained on text only. Second, we propose a new disentangled CLIP representation (DisCLIP) as standard CLIP embeddings are very entangled and have issues with attribute binding \cite{Ramesh2022_DALLE}. DisCLIP also integrates a syntactic parse of the text prompt.

\subsection{Focused Cross-Attention (FCA)}

To improve the binding of attributes to the correct objects, we restrict the attention of attributes to regions where their corresponding object has attention as well. Attribute dependencies are obtained from a dependency parse of the sentence and implemented in a binary matrix $D \in \{0,1\}^{n \times n}$, representing for each token of $y$ the token on which it is dependent. \footnote{The dependency matrix can implement complex relationships involving multiple objects and their respective attributes.}
Diffusion with FCA operates using two denoising model traversals, as formalized in Algorithm \ref{alg:FCA}. In the first traversal, standard cross-attention $A^l_t$ is used, from which the average attention maps $A^*$ are obtained by averaging $A^l_t$ over each layer $l$ and timestep $t$.
From these attention maps, we obtain the focus mask $F_\text{mask} \in \{- \infty, 0\}^{h \times w, n}$ and calculate FCA as follows:
\begin{equation}
\label{eq:fmask}
   F_\text{mask} = \delta (A^*D^{\textnormal{T}}), \; \text{with}
\end{equation}
\begin{equation}
\label{eq:delta}
   \delta(b_{ij}) = - \infty \text{ if }  \frac{b_{ij} - \underset{p}{\min}(b_{pj})}{\underset{p}{\max}(b_{pj}) -\underset{p}{\min}(b_{pj})} < s, \text{ else } 0
\end{equation}
\begin{equation}
\label{eq:FCA}
   \text{FCA}(Q, K, V, F_\text{mask}) = \text{softmax}(\dfrac{F_\text{mask} + QK^{\textnormal{T}}}{\sqrt{d}}) \; V
\end{equation}
where the threshold $s$ is a hyperparameter\footnote{More details about the selection of the threshold $s$ are presented in the Appendix \ref{influence_threshold_s}.}  and $\delta$ is a function that operates on each cell $b_{ij}$ of $A^*D^{\textnormal{T}}$. $F_\text{mask}= - \infty$ for the attributes' cross-attention map regions where its corresponding object has a normalized attention map value less than $s$. 
By replacing the cross-attention of $ \epsilon_\theta(x_t, y, t)$ with FCA (Equation \ref{eq:FCA}), we obtain $\epsilon_{\theta, FCA}(x_t, y, t, F_\text{mask})$. $\epsilon_{\theta, FCA}$ is then used to obtain the output image $\text{I}^*$ in a second model traversal with FCA.
Dimensions $w \times h$ are not the same for different layers $l$ in $\epsilon_\theta$. We use cubic interpolation to the largest layer size to average $A^l_t$ over layers of different sizes. Max pooling is used to project $F_\text{mask}$ back to the correct layer size of $l$.

A limitation of FCA is the double computational cost required by a second model run.\footnote{The computational cost of constructing and applying $F_\text{mask}$ is negligible compared to the diffusion cost.}
By stopping the first model run early, we reduce the increased cost without compromising the quality of the attention maps $A^*$. We obtain 99+\% of the performance on EPViT accuracy (on DAA-200 and CC-500) with only 10\% of the diffusion steps. 

 \begin{algorithm}[t]
 \scriptsize
\caption{Diffusion with FCA}\label{alg:FCA}
\hspace*{\algorithmicindent}\textbf{Input:} $\text{sentence encoding } y, \; \text{attribute dependencies } D$ \\
\begin{algorithmic}
\STATE $x_T \gets N (0, I)$
\FOR{$t \gets T...1$}
    \STATE $z_{t-1}, \{A_t^l\} \gets \epsilon_\theta(x_t, y, t)$
    \STATE $x_{t-1} \gets sample(x_t, z_{t-1})$
\ENDFOR
\STATE $A^* \gets \overline{A^l_t}$
\STATE $F_\text{mask} \gets \delta (A^*D^{\textnormal{T}})$ 
\STATE $x_T^* \gets x_T$
\FOR{$t \gets T...1$}
    \STATE $z_{t-1}^* \gets \epsilon_{\theta, FCA}(x_t^*, y, t, F_\text{mask})$
    \STATE $x_{t-1}^* \gets sample(x_t^*, z_{t-1}^*)$
\ENDFOR
\end{algorithmic}
\hspace*{\algorithmicindent}\textbf{Output:} $\text{image I}^* \gets x_0^*$
\end{algorithm}

\subsection{Disentangled CLIP Encoding (DisCLIP)}
\begin{figure}
    \begin{subfigure}{0.23\textwidth}
         \centering
        \includegraphics[width=\textwidth]{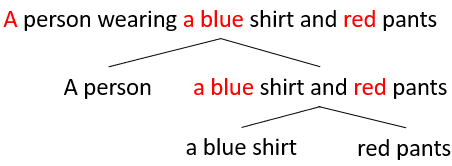}
         \caption{}
         \label{fig:const_tree}
     \end{subfigure}
     \hfill
    \begin{subfigure}{0.23\textwidth}
         \centering
        \includegraphics[width=\textwidth]{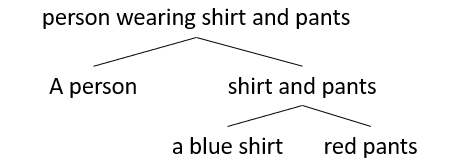}
          \caption{}
         \label{fig:abst_const_tree}
     \end{subfigure}
  \caption{An example of (a) a constituency tree and (b) an abstracted constituency tree (removed words in red).}
  \label{fig:tree}
\end{figure}

A CLIP encoding of a sentence consists of an embedding of each word concatenated with a sentence embedding and padding embeddings \cite{Radford2021clip}. T2I models based on CLIP encodings often have difficulties in image-text alignment \cite{saharia2022imagen}. 
We propose a novel training-free variation of CLIP, called DisCLIP. DisCLIP uses a syntactic parser to obtain a hierarchical representation of the text prompt in the form of a constituency tree. By replacing the noun phrases in the higher layers of the tree with their head nouns, we obtain an abstracted constituency tree (
Figure \ref{fig:tree}). The resulting tree encodes the compositional information including explicit object-attribute bindings.
The tree is then used to disentangle the CLIP representation of the text prompt. We independently encode the whole prompt or sentence, and each of the constituents of the tree with CLIP (padding embeddings removed) and concatenate the resulting embeddings. 
When used with FCA, an extra row and column are added to $D$ for each added constituent embedding, containing a dependency between the added constituent and the nouns that are present in it. The results below show that DisCLIP mitigates the problem of object-attribute binding.

Our DisCLIP encoding differs from the use of constituency trees in StructureDiffusion. First, we obtain one disentangled encoding of the text prompt in contrast to StructureDiffusion's multiple encodings.
This results in a more computational efficient diffusion process and allows integration with the proposed FCA method. Second, we introduce abstracted constituency trees that explicitly bind the local information of attributes to their corresponding objects.

\section{Experimental Set-up}
\subsection{Models}
We use open source T2I diffusion models and expand these with the FCA component and DisCLIP encoding of the text prompt. 
Experimental set-up and hyperparameter values are in Appendix \ref{exp_setup}. \\
{\bf Stable Diffusion SD} is trained on a filtered version of the LAION-5B \cite{schuhmann2022laion5b} dataset, uses the latent diffusion architecture of \citet{Rombach2022_ldm} and uses a frozen CLIP \cite{Radford2021clip} model as the text encoder.\\
{\bf DeepFloyd IF} is trained on the same data. Its design is based on the Imagen \cite{saharia2022imagen} architecture, which uses a frozen T5-XXL text encoder \cite{raffel2020t5} to generate images with a pixel-based diffusion process in three cascading stages, an initial image generation stage followed by two upsampling stages.\\
{\bf Attend-and Excite AE} refers to the original Attend-and-Excite model  \cite{Chefer2023AttentAndExcite}. It builds on Stable Diffusion and is designed to improve the generating of multiple objects mentioned in the text prompt by focusing the attention on nouns appearing in the text prompt.\\
{\bf Versatile Diffusion VD} extends an existing single flow diffusion pipeline into a multitask multimodal network that handles T2I, image-to-text and image-variation generation \cite{Xu_2023_ICCV}. \\
{\bf SynGen} \cite{rassin2023linguistic} 
relies on loss functions to align objects with their attributes. 
\\
{\bf Structure Diffusion} adapts the Stable Diffusion model to attend to multiple encodings of syntax constituents of the text prompt \cite{feng2023structguidance}.\\
The above models are used as baselines.\\
$\textbf{SD}_{\textbf{FCA}}, \textbf{SD}_{\textbf{DisCLIP}}$ and $\textbf{SD}_{\textbf{FCA+DisCLIP}}$ integrate FCA and DisCLIP into Stable Diffusion. \\
$\textbf{IF}_{\textbf{FCA}}$ represents DeepFloyd IF that integrates FCA. We did not integrate DisCLIP as the IF model does not support the CLIP embedding space.\\
$\textbf{AE}_{\textbf{FCA}}$ replaces the global attention of Attend-and-Excite computed at each timestep $t$ of the diffusion process by the proposed focused attention (FCA).
$\textbf{AE}_{\textbf{DisCLIP}}$ uses DisCLIP as encoder of the text prompt and $\textbf{AE}_{\textbf{FCA+DisCLIP}}$ integrates both.\\
$\textbf{VD}_{\textbf{FCA}}$ and $\textbf{VD}_{\textbf{DisCLIP}}$ represent the Versatile Diffusion model that integrates FCA and DisCLIP, respectively. $\textbf{VD}_{\textbf{FCA+DisCLIP}}$ integrates $\textbf{SynGen}_{\textbf{FCA}}$ and $\textbf{SynGen}_{\textbf{DisCLIP}}$ extend the SynGen model with FCA and DisCLIP, while $\textbf{SynGen}_{\textbf{FCA+DisCLIP}}$ integrates both.

\begin{table*}[t!]
    \centering
    \tiny
    \begin{tabular}{l cc cc cc}
    \toprule
     & \multicolumn{2}{c}{\text{DAA-200}}  & \multicolumn{2}{c}{\text{CC-500}} & \multicolumn{2}{c}{\text{AE-276}}  \\ 
    \cmidrule(l{2pt}r{2pt}){2-3}
    \cmidrule(l{2pt}r{2pt}){4-5}
    \cmidrule(l{2pt}r{2pt}){6-7}

    Methods & $\text{EPViT acc} \uparrow$ & $\text{CLIP score} \uparrow$ & $\text{EPViT acc} \uparrow$ & $\text{CLIP score} \uparrow$ 
    & $\text{EPViT acc} \uparrow$ & $\text{CLIP score} \uparrow$  \\

    \midrule
    Stable Diffusion SD & 59.1 $\pm$ 0.2 & 78.5 $\pm$ 0.1 & $61.1 \pm 0.3$ & \textbf{82.5 $\pm$ 0.1} & $47.5 \pm 4.6$ & \textbf{82.4 $\pm$ 0.4} \\
    Structure Diffusion & 62.3 $\pm$ 0.3 & \textbf{78.9 $\pm$ 0.2} & 61.1 $\pm$ 0.4 & 82.4 $\pm$ 0.3 & 40.8 $\pm$ 0.4 & 81.2 $\pm$ 0.5 \\
     $\textbf{SD}_{\textbf{FCA}} \textbf{ (ours)}$ & $60.4 \pm 0.1$ & 78.5 $\pm$ 0.1 & $62.1 \pm 0.2$ & \textbf{82.5 $\pm$ 0.2} & \textbf{46.8 $\pm$ 5.3} & \textbf{82.4 $\pm$ 0.4} \\
     $\textbf{SD}_{\textbf{DisCLIP}} \textbf{ (ours)}$ & $61.4 \pm 0.5$ & $78.2 \pm 0.1$ & $62.8 \pm 0.5$ & $81.4 \pm 0.1$ & $44.7 \pm 3.8$ & $81.1 \pm 0.1$ \\
     $\textbf{SD}_{\textbf{FCA+DisCLIP}} \textbf{ (ours)}$ & \textbf{63.5 $\pm$ 0.5} & $78.4 \pm 0.1$ & \textbf{66.3 $\pm$ 10} & $82.0 \pm 0.2$ & \textbf{47.8 $\pm$ 2.3} & $82.3 \pm 0.1$ \\
     \midrule
     \text{DeepFloyd IF} & $69.3 \pm 0.0$ & \textbf{78.3 $\pm$ 1.4} & $64.4 \pm 5$ & \textbf{85.3 $\pm$ 1.9} & $45.5 \pm 0$ & $86.0 \pm 0.1$ \\
     $\textbf{IF}_{\textbf{FCA}} \textbf{ (ours)}$ & \textbf{72.8 $\pm$ 0.3} & $78.2 \pm 1.4$ & \textbf{70.4 $\pm$ 2.1} & $84.7 \pm 2.4$ & \textbf{48.5 $\pm$ 3.0} & \textbf{86.1 $\pm$ 0.0} \\
     \midrule
     \text{Attend-and-Excite AE}& $73.4 \pm 0.6$ & $79.5 \pm 0$ & $74.4 \pm 0.4$ & \textbf{87.1 $\pm$ 0.1} & $56.1 \pm 3.1$ & \textbf{86.3 $\pm$ 0.3} \\
     $\textbf{AE}_{\textbf{FCA}} \textbf{ (ours)}$& $73.2 \pm 0.3$ & $79.4 \pm 0.0$ & $74.9 \pm 0.4$ & $86.4 \pm 0.1$ & $52.3 \pm 0.8$ & \textbf{86.3 $\pm$ 0.0} \\
     $\textbf{AE}_{\textbf{DisCLIP}} \textbf{ (ours)}$& $73.5 \pm 0.1$ & $79.5 \pm 0.0$ & $74.6 \pm 0.8$ & $87.0 \pm 0.1$ & $55.3 \pm 0.8$ & \textbf{86.3 $\pm$ 0.1} \\
     $\textbf{AE}_{\textbf{FCA+DisCLIP}} \textbf{ (ours)}$& \textbf{73.5 $\pm$ 0.6} & \textbf{79.6 $\pm$ 0.1} & \textbf{75.0 $\pm$ 0.5} & \textbf{87.1 $\pm$ 0.2} & \textbf{56.5 $\pm$ 3.1} & \textbf{86.3 $\pm$ 0.1} \\
     \midrule
     \text{Versatile Diffusion VD}& 54.0 $\pm$ 0.5 & 77.4 $\pm$ 0.1 & $52.4 \pm 0.0$ & $81.2 \pm 0.4$ & $28.8 \pm 1.5$ & $79.8 \pm 0.0$ \\
     $\textbf{VD}_{\textbf{FCA}} \textbf{ (ours)}$& \textbf{54.5 $\pm$ 1.0} & \textbf{77.5 $\pm$ 0.2} & $53.5 \pm 0.4$ & $81.3 \pm 0.4$ & $31.1 \pm 0.8$ & \textbf{79.9 $\pm$ 0.0} \\
     $\textbf{VD}_{\textbf{DisCLIP}} \textbf{ (ours)}$& $53.0 \pm 1.5$ & $77.1 \pm 0.2$ & \textbf{55.6 $\pm$ 0.8} & 80.6 $\pm$ 0.5 & $28.8 \pm 1.5$ & $78.4 \pm 0.4$ \\
     $\textbf{VD}_{\textbf{FCA+DisCLIP}} \textbf{ (ours)}$& 54.0 $\pm$ 0.8 & 77.4 $\pm$ 0.2 & \textbf{55.6 $\pm$ 0.8} & \textbf{80.8 $\pm$ 0.8} & \textbf{33.3 $\pm$ 1.5} & $78.4 \pm 0.6$ \\
     \hline
     \text{SynGen}& 80.3 $\pm$ 0.9 & 79.2 $\pm$ 1.5 & $87.8 \pm 0.8 $ & $87.0 \pm 0.5$ & \textbf{63.8 $\pm$ 1.8} & \textbf{84.5 $\pm$ 0.08}\\
     $\textbf{SynGen}_{\textbf{FCA}} \textbf{ (ours)}$& 82.3 $\pm$ 1.0 & \textbf{79.3 $\pm$ 0.4} & $87.9 \pm 0.9$ & $86.9 \pm 0.7$ & $63.2 \pm 0.6$ & 84.3 $\pm$ 0.5 \\
     $\textbf{SynGen}_{\textbf{DisCLIP}} \textbf{ (ours)}$& $86.5 \pm 0.2$ & $78.9 \pm 0.5$ & 88.0 $\pm$ 0.4 & \textbf{87.3 $\pm$ 0.4} & $62.3 \pm 1.6$ & $84.1 \pm 0.7$ \\
     $\textbf{SynGen}_{\textbf{FCA+DisCLIP}} \textbf{ (ours)}$ & \textbf{86.7 $\pm$ 1.1} & 78.4 $\pm$ 0.3 & \textbf{88.1 $\pm$ 0.9} & 87.1 $\pm$ 0.5 & 63.2 $\pm$ 1.6 & $83.9 \pm 0.7$ \\
    \bottomrule
    \end{tabular}
    \caption{
    Evaluation of the enhanced models with the FCA and DisCLIP components compared to baseline models. The evaluation relies on EPViT accuracy and CLIP score (mean and variance).
    }
    \label{tab:cleaned_results}
\end{table*}

\subsection {Datasets and Metrics } 

\subsubsection {Evaluation of EPViT} We evaluate the newly introduced EPViT accuracy on the DAA-200 dataset implementing a classification set-up (described in Section 4.1) and compare results with those obtained with the CLIP score.

\subsubsection{Evaluation of the T2I generation} 
We evaluate the object-attribute binding of the models on dedicated datasets and their general T2I generation capabilities on a general-purpose dataset.\\
{\bf Evaluation of object-attribute binding}
We report results obtained on the DAA-200 (see Section 4.1), Concept Conjunction 500 (CC-500) \cite{feng2023structguidance} and the Attend-and-Excite (AE-276) \cite{Chefer2023AttentAndExcite} datasets. 
Results are quantitatively evaluated using the EPViT accuracy and CLIP score (ViT-L/14) \cite{Radford2021clip}.
We use two human evaluations to assess the image fidelity and image-text alignment. First, we ask annotators to compare two generated
images and indicate which image demonstrates better image-text alignment and image fidelity. 
Second, following \citet{feng2023structguidance}, we ask annotators whether the two objects of the CC-500 samples are present in the generated images and whether they are in the correct color. We also ask whether a part of the object is in the color of the other object to assess how many attributes are leaked to the wrong objects. 
Results are shown in Tables \ref{tab:cleaned_results}-\ref{tab:obj_results}. Details on the human evaluation are in Appendix \ref{detail_human_evaluation}.\\
{\bf Evaluation of general T2I generation capabilities}
To evaluate the T2I performance on more general prompts that refer to multiple attributes and objects - up to 14, we evaluate on 10K randomly sampled captions from MSCOCO (COCO-10K) \cite{Lin2014mscoco}.
To assess image fidelity, we
follow previous work and report the Inception Score (IS) \cite{Salimans2016IS} and Frechet Inception Distance (FID) \cite{Heusel2017FID} apart from the CLIP score.
Results are 
in Table \ref{tab:general_prompts}.


\section{Results and Discussion}
\subsection{Evaluation of EPViT}
To evaluate the proposed EPViT model on the DAA-200 evaluation benchmark, we assess its classification accuracy on the ground truth images of DAA-200 and report the accuracy of correctly classifying the ground truth sentences above the adversarial ones. We compare the accuracy of EPViT against the classification accuracy using the CLIP score (details are in Appendix \ref{std_dev_epvit}). 
EPViT achieves an accuracy of 91\% and outperforms the CLIP score (63\%) by 28\%. 
The poor performance of CLIP aligns with 
\cite{saharia2022imagen, Ramesh2022_DALLE} showing that CLIP has issues with the correct binding of attributes to objects, which might be due to
the contrastive pretraining of CLIP, in which a complete caption is paired with a full image.
We conclude that the EPViT accuracy is superior to the CLIP score for the evaluation of the image-text alignment in the DAA-200 benchmark. 

\begin{table}[t!]
    \centering
    \tiny
    \begin{tabular}{l c cc cc}
    \toprule
      & \textbf{ours} & \multicolumn{2}{c}{\text{Alignment}} & \multicolumn{2}{c}{\text{Fidelity}} \\ 
    \cmidrule(l{2pt}r{2pt}){3-4} 
    \cmidrule(l{2pt}r{2pt}){5-6}
      \textbf{Benchmark} & \textbf{v.s.} & $\text{Win}\uparrow$ & $\text{Lose}\downarrow$ & $\text{Win}\uparrow$ & $\text{Lose}\downarrow$ \\
     \midrule
      \multirow{3}{*}{\textbf{DAA-200}} & StructureDiffusion & \textbf{36.1} & 34.0  & \textbf{37.0} & 32.9 \\
    &DeepFloyd IF & \textbf{22.3} & 17.8 & \textbf{29.6} & 27.9 \\
    & Attend-and-Excite  & \textbf{32.5} & 28.5 & 37.0 & \textbf{38} \\
    & Versatile Diffusion  & \textbf{34.2} & 31.3 & \textbf{37.3} & 33.8 \\
    & SynGen & \textbf{33.2} & 30.3 & \textbf{36.2} & 32.7 \\
    \midrule
     \multirow{3}{*}{\textbf{CC-500}} & StructureDiffusion & \textbf{32.4} & 28.2 & \textbf{43.3} & 28.4\\
    & DeepFloyd IF & \textbf{27.3} & 13.3 & \textbf{30.0} & 22.1\\
    & Attend-and-Excite  & \textbf{49.7} & 24.5 & \textbf{45.8} & 31.8 \\
    & Versatile Diffusion  & \textbf{28.1} & 27.7 & \textbf{37.5} & 33.1 \\
    & SynGen & \textbf{36.5} & 34.4 & \textbf{39.7} & 34.7 \\
    \midrule
    \multirow{3}{*}{\textbf{AE-276}} & StructureDiffusion & \textbf{36.2} & 27.8 & \textbf{45.6} & 33.3\\
    & DeepFloyd IF & \textbf{31.5} & 28.6 & \textbf{43.1} & 33.6 \\  
    & Attend-and-Excite  & \textbf{36.3} & 27.1 & \textbf{39.8} & 36.9 \\
    & Versatile Diffusion  & \textbf{29.3} & 27.3 & 29.2 & \textbf{30.7}\\
    & SynGen & 31.3 & \textbf{34.2} & 34.6 & \textbf{35.1} \\
    \bottomrule
    \end{tabular}
    \caption{Percentage of cases in which our FCA and DisCLIP modules generate better (win) or worse (lose) alignment and T2I fidelity than their baselines. 
    }
    \label{tab:compare_results}
\end{table}

\begin{table}[t!]
    \centering
    \tiny
    \begin{tabular}{l ccc}
    \toprule
      \text{Methods} & $\text{Two objects}\uparrow$  &  $\text{Atleast one object}\uparrow$  &  $\text{Leakage}\downarrow$  \\
    \midrule
     \text{Stable Diffusion} & 20.7 & 76.9 & 64.9 \\
    \text{StructureDiffusion} & 21.2 & 77.2 & 63.9 \\
    $\textbf{SD}_{\textbf{DisCLIP+FCA}} \textbf{ (ours)}$ & \textbf{22.2} & \textbf{80.8} & \textbf{56.8} \\
    \midrule
    \text{DeepFloyd IF} & 35.6 & 74.6 & 73.0 \\
    $\textbf{IF}_{\textbf{FCA}} \textbf{ (ours)}$ & \textbf{43.1} & \textbf{82.6} & \textbf{60.6} \\
    \midrule
    \text{Attend-and-Excite AE} & 46.8 & 88.4 & 65.4 \\
    $\textbf{AE}_{\textbf{FCA+DisCLIP}} \textbf{ (ours)}$ & \textbf{60.2} & \textbf{94.3} & \textbf{64.5} \\    
    \midrule
    \text{Versatile Diffusion VD} & 23.4 & 72.8 & 77.5 \\
    $\textbf{VD}_{\textbf{FCA+DisCLIP}} \textbf{ (ours)}$ & \textbf{25.6} & \textbf{76.3} & \textbf{69.7} \\  

    \midrule
    \text{SynGen} & 45.3 & 90.3 & 32.1 \\
    $\textbf{SynGen}_{\textbf{FCA+DisCLIP}} \textbf{ (ours)}$ & \textbf{47.2} & \textbf{91.2} & \textbf{27.4} \\  
    \bottomrule
    \end{tabular}
    \caption{Results of the human evaluation obtained on CC-500. We show how often (in \%) each model is able to correctly (with the correct color) generate at least one object / the two objects of the CC-500 captions. Leakage displays how often (in \%) an object is at least partially generated with the color of the wrong object.}
    \label{tab:obj_results}
\end{table}

\begin{table}[t!]
    \centering
    \tiny
    \begin{tabular}{l ccc}
    \toprule
    Methods & $\text{CLIP score} \uparrow$ & $\text{FID score} \downarrow$ & $\text{IS acc} \uparrow$  \\

    \midrule
    \text{Stable Diffusion SD} & \textbf{79.9 $\pm$ 0.4} & \textbf{18.1 $\pm$ 0.4} & $37.7 \pm 0.2$ \\
     $\textbf{SD}_{\textbf{FCA+DisCLIP}}$  & $78.4 \pm  0.5$ & \textbf{18.1 $\pm$ 0.2} & \textbf{38.0 $\pm$ 0.2} \\
     \midrule
     \text{DeepFloyd IF}  & \textbf{78.3 $\pm$ 0.1} & \textbf{53.6 $\pm$ 0.1} & \textbf{27.7 $\pm$ 0} \\
     $\textbf{IF}_{\textbf{FCA}}$  & $78.2 \pm 0.1$ & \textbf{53.6 $\pm$ 0.1} & \textbf{27.7 $\pm$ 0} \\
     \midrule
     \text{Attend-and-Excite AE}  & $79.5 \pm 0.3$ & $18.2 \pm 0.5$ & $35.0 \pm 0.1$\\

     $\textbf{AE}_{\textbf{FCA+DisCLIP}}$  & \textbf{79.6 $\pm$ 0.4} & \textbf{18.1 $\pm$ 0.2} & \textbf{35.2 $\pm$ 0.2}\\
     \midrule
     \text{Versatile Diffusion VD} & \textbf{79.0 $\pm$ 0.3} & 26.3 $\pm$ 0.1 & 34.1 $\pm$ 0.0\\
     $\textbf{VD}_{\textbf{FCA+DisCLIP}}$ & 78.3 $\pm$ 0.5 & \textbf{25.2 $\pm$ 0.5} & \textbf{34.2 $\pm$ 0.1}\\

     \midrule
     \text{SynGen} & \textbf{87.9 $\pm$ 0.2} & 15.3 $\pm$ 0.3 & 36.7 $\pm$ 0.2\\
     $\textbf{VD}_{\textbf{FCA+DisCLIP}}$ & 87.3 $\pm$ 0.4 & \textbf{15.1 $\pm$ 0.5} & \textbf{36.9 $\pm$ 0.2}\\
    \bottomrule
    \end{tabular}
    \caption{Results of the FCA and DisCLIP enhanced models and their baselines obtained on the general prompts of COCO 10-K in terms of CLIP, FID and IS scores (mean and variance).
    }
    \label{tab:general_prompts}
\end{table}
\begin{figure*}[t]
    \centering
   \includegraphics[width=0.95\textwidth]{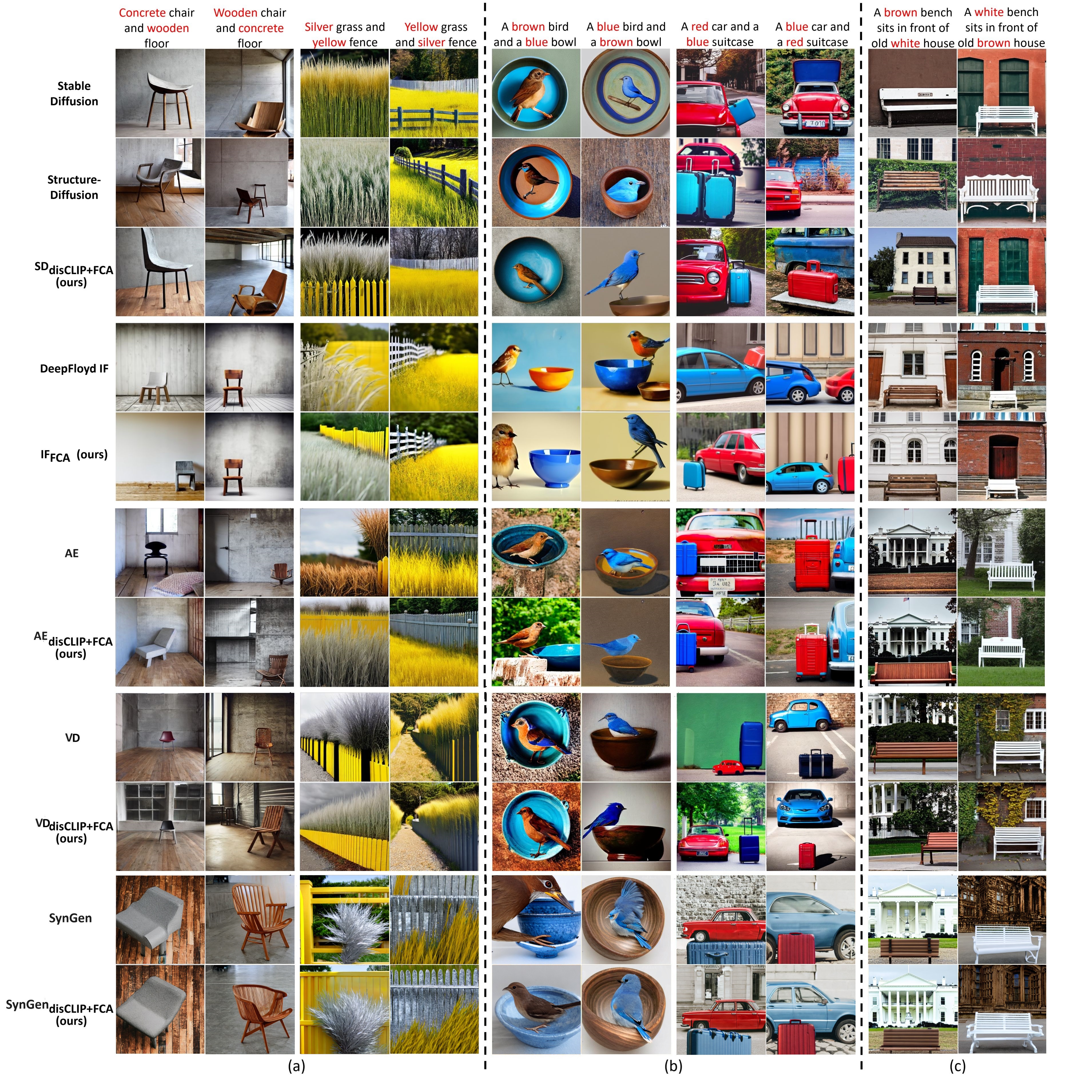}
    \caption{Qualitative results that show that the FCA and DisCLIP enhanced models improve attribute binding and decrease attribute leakage in images from (a) DAA-200, (b) CC-500 and (c) COCO-10K.}
    \label{fig:images}
\end{figure*}

\subsection{Evaluation of T2I Generation}
\paragraph{Evaluation of object-attribute binding}
In this section, we discuss the results obtained on datasets that challenge object-attribute binding, which are DAA-200, CC-500 and AE-276 datasets. The results expressed as EPViT and CLIP scores are found in Table \ref{tab:cleaned_results}. The EPViT scores of the enhanced models with FCA and DisCLIP indicate that the binding between the objects and their attributes is always better (up to 5 percentage points) than the binding achieved by the baseline models. 
Solely integrating FCA or DisCLIP already results in an improved object-attribute binding 
as the EPViT scores are higher (in a few cases on par) than the scores of the baseline models. As expected, the CLIP scores are very similar: They compute the semantic match between the full image and the text prompt neglecting syntactic dependencies of the  prompt and corresponding object-attribute binding. 

Table \ref{tab:compare_results} presents the results of the human evaluation. Observe that for DAA-200, CC-500 and AE-276 our methods outperform the baselines considering image-text alignment and all baselines but one with regard to image fidelity. 
The largest increase is seen on CC-500 where our methods outperform other models by 4-25 percentage points on alignment and 8-15 percentage points on fidelity. We hypothesize that T2I generation struggles when it is not straightforward which attribute belongs to which object. Because color attributes often co-occur with many objects, the captions of CC-500 are especially difficult for T2I models (as they contain only color attributes). This leads to a larger improvement for our models that explicitly bind attributes to certain regions, as can be seen in Figure \ref{fig:images}b. DAA-200 and AE-267, on the other hand, contain diverse categories of attributes. Base models are good at generating the most expected attribute binding. Unlike our models, they perform poorly when attributes are switched. An example is shown in Figure \ref{fig:images}a where all models perform well on the prompt ``yellow grass and silver fence'' but only our models perform well on the prompt ``Silver grass and yellow fence''. Although SynGen already achieves accurate object-attribute alignments, 
when augmenting this model with FCA and DisCLIP, the image quality is enhanced the and leakage is decreased.
More qualitative examples are in Appendix \ref{additional_results}. 
Table \ref{tab:obj_results} successfully evaluates object-attribute binding of the proposed methods by conducting a human evaluation that checks whether each object of CC-500 is generated with the correct color. We perform this analysis on CC-500 as the attributes are only colors that are associated with a wide range of objects. 
$\text{SD}_{\text{DisCLIP+FCA}}$ and $\text{IF}_{\text{FCA}}$ decrease leakage by 8 and 2.5 percentage points, respectively. 
Appendix \ref{EPViT_human_evaluation} shows that EPVit aligns with human evaluation based on a sample-by-sample comparison.

\subsection{Evaluation of General Prompts}
Our models do not lose any quality when conditioned on the general prompts of COCO-10K, evidenced by the FID and IS scores in Table \ref{tab:general_prompts}. 

\section{Conclusion}

We have proposed training-free methods 
to emphasize the importance of integrating linguistic syntactic structures in T2I generation. 
We 
demonstrated their easy and successful integration in state-of-the-art T2I diffusion models leading to an improved object-attribute binding and to a decrease in attribute leakage in the generated image. Additionally, we have shown that a newly designed metric, EPViT, outperforms CLIP in assessing object-attribute binding of T2I models. EPViT allows for a better understanding and measurement of a model's performance in accurately reflecting the intended textual description in the generated image.

\section*{Limitations}
Our proposed methods can be applied across various image generation models and demonstrate broad versatility and wide applicability. This contributes greatly to the field by offering a flexible solution that can improve existing models and potentially accelerate the development of new ones. However, our methods have some limitations. Currently, the training of the EPViT model is focused on object-attribute binding. Similarly, the proposed focused cross-attention (FCA) is designed for object-attribute binding, while the abstracted constituency tree leveraged for DisCLIP accomplishes the abstraction of noun phrases. Future research can enlarge the scope of this work by considering other syntactic relationships found in text prompts to improve T2I generation. Moreover, the proposed methods of this paper are restricted by the expressiveness and correctness of the syntactic parses and their ability to deal with linguistic phenomena such as shared modifiers, choice of head instead of multiple modifiers, coordinating conjunctions, and other complex syntactic structures found in the text prompt, a problem which can be aggregated when dealing with languages that have limited syntactic parsers. We leave it to future work to investigate their effect on T2I generation.

We did not encounter severe limitations with regard to scalability of the models and the necessary GPU resources. 

\section*{Ethics Statement}

T2I generation helps users to automatically create images given a textual prompt. This technology also facilitates the creation of fake content. The latter often involves the generation of public figures in fake situations unlike our work which focuses on improving the generation of objects and their attributes. Nevertheless, our contributions could be misused for the creation of misleading content.

The annotators involved in the human evaluation were paid through the Amazon MTurk platform within three working days after completion. They earned \$18/hour, which is superior to the minimum wage. We did not discriminate between the annotators in terms of gender, race, religion, or any other demographic feature.


\bibliography{custom}
\bibliographystyle{acl_natbib}

\appendix

\section{Appendix}
\label{sec:appendix}

\subsection{Diffusion Models}\label{diffusion}

Diffusion models \cite{saharia2022imagen, Nichol2021Glide} are probabilistic generative models that generate images $x_0$ with width $w$ and height $h$ by progressively removing noise from an initial vector of Gaussian noise $x_T$.
These models accomplish this by training a denoising network $\epsilon_\theta(x_t, t)$ that for each timestep $t$ and noised image $x_t =  \sqrt{\alpha_t} x_0 + \sqrt{1 - \alpha_t} z$ estimates the added noise $z$, where $z \sim N(0, I)$ and $\{\alpha_t\}$ represents the noise schedule.  Once trained, inference consists of gradually denoising $x_T$ by iteratively applying $\epsilon_\theta$ to the current $x_t$ and adding a Gaussian noise perturbation to obtain a cleaner $x_{t-1}$.
In T2I generation, the denoising network $\epsilon_\theta(x_t, y, t)$ 
is conditioned on textual input $y \in\mathbb{R}^{n, d_y}$ 
containing $n$ token embeddings of dimension $d_y$.  $\epsilon_\theta$ is commonly implemented with a U-Net architecture \cite{ronneberger2015unet} which 
consists of multiple U-Net layers $l$ using cross-attention to condition the generation on $y$.
\citet{Rombach2022_ldm} introduced diffusion in a latent space instead of the pixel space
hereby substantially reducing model size. They use a pretrained variational autoencoder to encode the images into a smaller latent space. This results in a diffusion process where $x_t$ is a latent encoding of an image instead of the image itself. At inference, an image is obtained by decoding the denoised latent $x_0$ with the decoder of the autoencoder.

\subsection{Details of the Experimental Setup and Hyperparameter Values}\label{exp_setup}

 EPViT is initialized from CLIP's ViT-L/14. FCA uses a threshold $s$ of $0.5$. EPViT was trained with a batch size of 128 on $4 \times 24$ GB GPUs (NVIDIA GeForce RTX 3090) for 8 epochs of the filtered Visual Genome dataset.
 
 In our experiments, all Stable Diffusion evaluations are executed on an image size of 512x512 with Stable Diffusion v1.5. All DeepFloyd IF runs are evaluated on an image size of 256x256 using the first 2 stages of DeepFloyd IF large v1.0. All comparisons use the same seeds for each model with 50 diffusion steps and a guidance scale of 7.5. For the Stable Diffusion runs, we use the LMSDiscrete scheduler \cite{karras2022shedulers} for sampling; for the DeepFloyd IF runs, the discrete denoising scheduler \cite{Ho2020DDPM} is used. Dependency and constituency parses are obtained with the LAL-parser of \citet{Mrini2019lalparser}. 

Stable Diffusion runs were executed on NVIDIA GeForce RTX 3080 GPUs, requiring 12GB of GPU RAM; DeepFloyd IF runs on NVIDIA RTX 3090 GPUs, requiring 19GB of GPU RAM. 16bit precision was used for all runs.

When quantitatively evaluating the T2I generation,  for DAA-200, we generate 10 images per prompt; for CC-500 we follow \citet{feng2023structguidance} and generate 3 images per prompt; for AE-276 we follow \cite{Chefer2023AttentAndExcite} and generate one image per prompt.
Because CC-500 contains only constructed sentences, we 
automatically generate the graphs needed for EPViT. 
Baselines and their extensions with our proposed methods use the same initialization of parameters. We used three seeds for each prompt.



\subsection{Details of the Human Evaluation}\label{detail_human_evaluation}

Our human evaluation consists of a comparison and object detection tasks. In our comparison tasks we compare two T2I models by asking annotators, given a prompt and a generated image for each model, which image has the best (or equal) image-text alignment and which has the best (or equal) image fidelity. To assess the image-text alignment, we asked annotators ``Which image displays best what is mentioned in the text?''; to assess image fidelity, we asked ``Regardless of the text, which image is more realistic and natural?''. We randomly swap images around to prevent order bias from annotators.

For our object detection task we ask annotators for each generated image of CC-500 six multiple choice questions about objects in the images. Note that these images are generated from prompts of the form ``a $\langle color1\rangle \langle object1 \rangle$ and a $\langle color2 \rangle \langle object2\rangle$''. We ask ``is there a $\langle object1 \rangle$'' and if answered yes we follow this with the questions: ``is most of the $\langle object1 \rangle \langle color1 \rangle$'' and ``is some part of the $\langle object1 \rangle \langle color2 \rangle$''. We ask the same questions for $\langle object2 \rangle$. We asked the same annotator to evaluate all models per prompt to limit variability which can occur because of different interpretations, between annotators, of how an object should look like to be classified as an object.

The human evaluation was executed with the use of the Amazon Mechanical Turk platform. For both kinds of evaluation tasks we group our samples in batches of HITs. We select workers from the United States and pay for each HIT 0.15 or 0.20 US dollar, for the comparison and object detection tasks. Comparison HITs take on average approximately 20-40 seconds to complete and object detection HITs take approximately 30-50 seconds. This results in an average hourly payment of 18 US dollars. On each comparison task worked between 28 and 36 different annotators; On each object detection task worked 54 different annotators. 

{\subsection{Correlation Between Human Evaluation and EPViT}
\label{EPViT_human_evaluation}
When we evaluate the sample-by-sample correlation between EPViT and the alignment metric (evaluated by the annotators and presented in Table 2) the obtained score is 0.59, indicating a satisfactory correlation. EPViT is also well correlated with the correct object-attribute binding evaluated by the annotators in Table 3. When at least one object is correctly connected to its attribute, the sample-by-sample correlation with EPViT is 0.73. When two objects are correctly connected with their attribute the sample-by-sample correlation with EPViT is 0.51. To compute correlation scores we used the Pearson coefficient and discarded the cases where the annotators did not make a decision. The good correlation with human evaluation indicates that EPViT can successfully evaluate the aspect that methods such as FCA, DisCLIP and SynGen are designed for. 


\subsection{Ablation Study}
\subsubsection{Standard Deviation of the EPViT Accuracy}\label{std_dev_epvit}
Evaluations with the EPViT accuracy do not depend on randomness, however the generation of the images does. In this work we calculate the EPViT accuracy on the same set of images used in the human evaluation, as this allows for the most fair comparison between the metrics and the human evaluation. To test the variability of the EPViT accuracy we generate five times with different seeds 10 images for each prompt of DAA-200 (resulting in 2000 images per evaluation). The resulting EPViT accuracies have a standard deviation in between $0.4$ and $0.7$ as can be seen in Table \ref{tab:std}. The standard deviation could be further reduced by evaluating the EPViT accuracy over a bigger set of generated images.

\begin{table}[ht]
    \centering
    \tiny
    \begin{tabular}{l cccc}
    \toprule
      EPViT accuracy &Stable Diffusion  &  $\text{SD}_{\text{DisCLIP+FCA}}$ & DeepFloyd IF & $\text{IF}_{\text{FCA}}$  \\
    \midrule
     \text{average} & 59.2 & 64.2 & 68.6 & 72.9 \\
    \text{standard deviation} & 0.6 & 0.4 & 0.6 & 0.7 \\
    \end{tabular}
    \caption{The average and standard deviation of the EPViT accuracy calculated from five sets of generated images per model on the DAA-200 dataset.}
    \label{tab:std}
\end{table}

\subsubsection{Influence of Threshold $s$} \label{influence_threshold_s}

\begin{figure}
\includegraphics[width=0.48\textwidth]{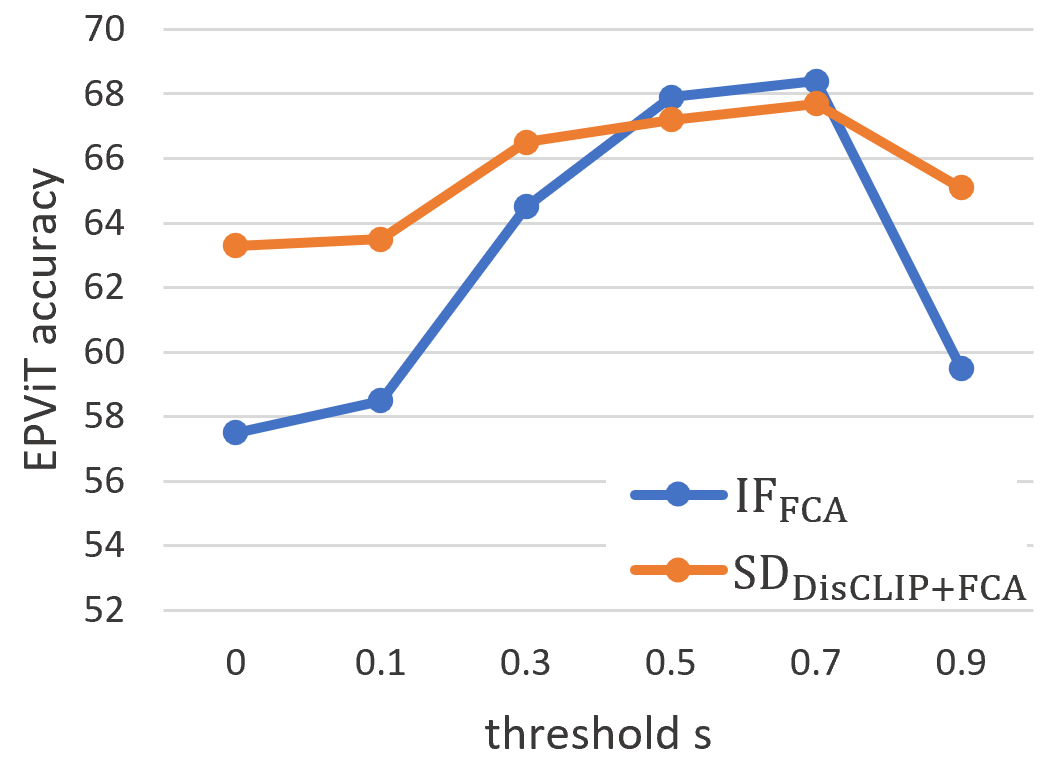}
  \caption{ (a) The classification accuracy in \% on the ground truth images of DAA-200. (b) The influence of different threshold values $s$ on the EPViT accuracy (in \%) on CC-500.}
  \label{fig:abl_threhold}
\end{figure}

The results of using different threshold values $s$ when computing FCA are visualized in Figure \ref{fig:abl_threhold} and show the best performance when $s$ is between 0.5 - 0.7. A high threshold value yields degenerated results because FCA masks the attributes over too many regions of the image. In contrast, when $s$ is too low the attention is unrestricted, yielding results that are identical to those of the base model.\\

\subsubsection{Visualization of Attention Maps}
We calculate the averaged cross-attention maps over all steps and layers of the diffusion process. In Figure \ref{fig:app:if_att} you can see that without the use of FCA the attention over attributes is spread across the whole image. Using FCA restricts the attention of attributes to the same spatial locations as their corresponding objects, resulting in improved attribute binding.


\begin{figure*}[ht]
    \centering
   \includegraphics[width=0.5\textwidth]{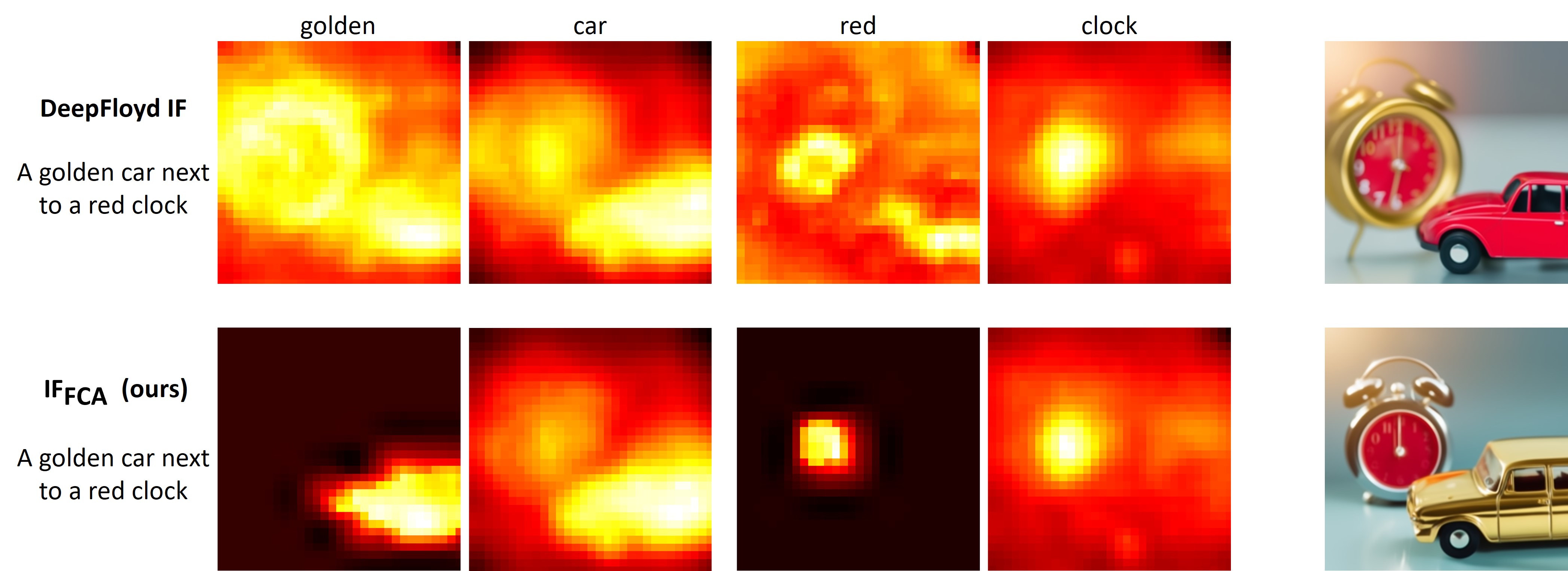}
    \caption{Visualization of the influence of FCA on the averaged cross-attention maps of DeepFloyd IF. We compare the attention maps of attribute and object tokens.}
    \label{fig:app:if_att}
\end{figure*}


\subsection{Additional Results}\label{additional_results}
Figure \ref{fig:app:sd_results} and \ref{fig:app:if_results} show additional qualitative results of using FCA on top of Stable Diffusion and DeepFloyd IF, respectively.

\begin{figure*}[ht]
    \centering
    \includegraphics[width=0.9\textwidth]{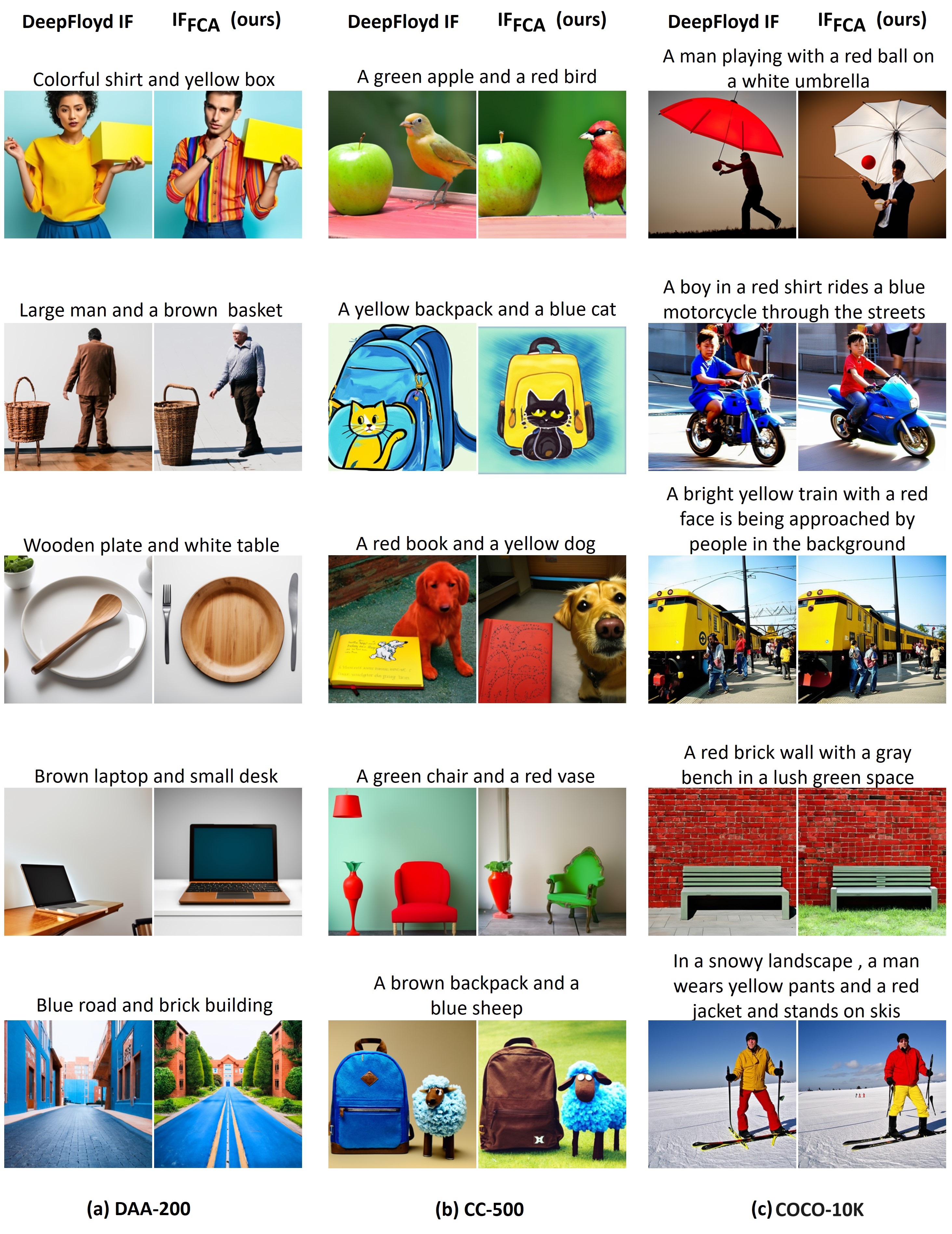}
    \caption{Qualitative results showing the influence of FCA on DeepFloyd IF. Examples from the datasets: (a) DAA-200, (b) CC-500 and (c) COCO-10K.}
    \label{fig:app:if_results}
\end{figure*}

\begin{figure*}[ht]
    \centering
    \includegraphics[width=0.9\textwidth]{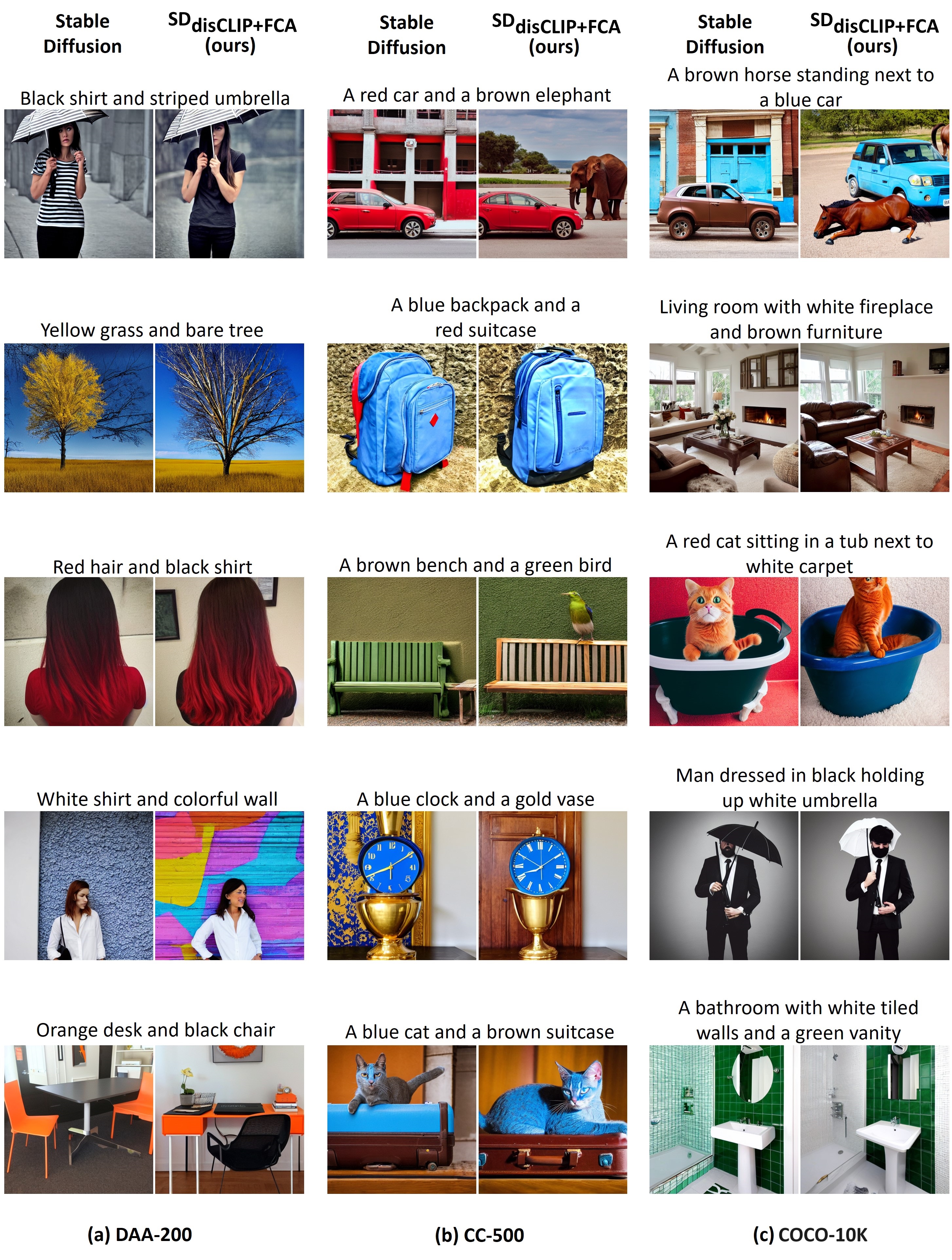}
    \caption{Qualitative results showing the influence of DisCLIP and FCA on Stable Diffusion. Examples from the datasets: (a) DAA-200, (b) CC-500 and (c) COCO-10K.}
    \label{fig:app:sd_results}
\end{figure*}



\begin{figure*}[ht]
    \centering
    \includegraphics[width=0.9\textwidth]{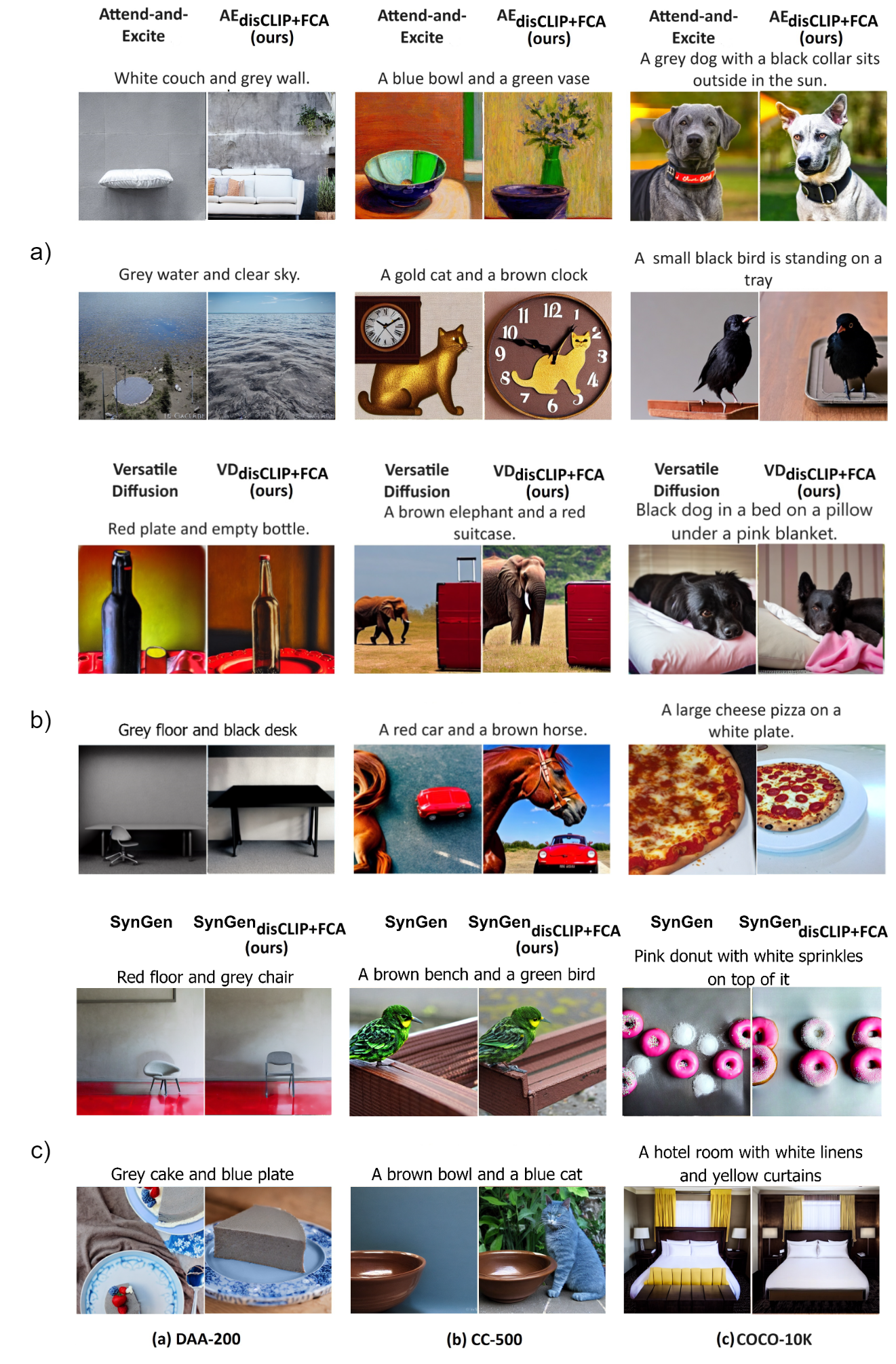}
    \caption{Qualitative results showing the influence of DisCLIP and FCA on the Attend-and-Excite (a), Versatile Diffusion (b) and SynGen (c) models. Examples from the datasets: (a) DAA-200, (b) CC-500 and (c) COCO-10K.}
    \label{fig:app:syngen_results}
\end{figure*}
\end{document}